\PassOptionsToPackage{unicode}{hyperref}
\PassOptionsToPackage{hyphens}{url}
\PassOptionsToPackage{dvipsnames,svgnames,x11names}{xcolor}
\documentclass[
  american,
  11pt,
  letterpaper,
]{article}
\usepackage{xcolor}
\usepackage[margin=1in]{geometry}
\usepackage{amsmath,amssymb}
\usepackage{iftex}
\ifPDFTeX
  \usepackage[T1]{fontenc}
  \usepackage[utf8]{inputenc}
  \usepackage{textcomp} 
\else 
  \usepackage{unicode-math} 
  \defaultfontfeatures{Scale=MatchLowercase}
  \defaultfontfeatures[\rmfamily]{Ligatures=TeX,Scale=1}
\fi
\usepackage{lmodern}
\ifPDFTeX\else
\fi
\IfFileExists{upquote.sty}{\usepackage{upquote}}{}
\IfFileExists{microtype.sty}{
  \usepackage[]{microtype}
  \UseMicrotypeSet[protrusion]{basicmath} 
}{}
\makeatletter
\@ifundefined{KOMAClassName}{
  \IfFileExists{parskip.sty}{%
    \usepackage{parskip}
  }{
    \setlength{\parindent}{0pt}
    \setlength{\parskip}{6pt plus 2pt minus 1pt}}
}{
  \KOMAoptions{parskip=half}}
\makeatother
\usepackage{longtable,booktabs,array}
\usepackage{caption}
\usepackage{calc} 
\usepackage{etoolbox}
\makeatletter
\patchcmd\longtable{\par}{\if@noskipsec\mbox{}\fi\par}{}{}
\makeatother
\IfFileExists{footnotehyper.sty}{\usepackage{footnotehyper}}{\usepackage{footnote}}
\makesavenoteenv{longtable}
\ifLuaTeX
\usepackage[bidi=basic,shorthands=off]{babel}
\else
\usepackage[bidi=default,shorthands=off]{babel}
\fi
\ifLuaTeX
  \usepackage{selnolig} 
\fi
\providecommand{\tightlist}{%
  \setlength{\itemsep}{0pt}\setlength{\parskip}{0pt}}
\usepackage{graphicx}
\usepackage{microtype}
\usepackage{xurl}
\usepackage{bookmark}
\IfFileExists{xurl.sty}{\usepackage{xurl}}{} 
\makeatletter
\@ifundefined{xmpquote}{}{}
\makeatother
\hypersetup{
  pdftitle={The Price of Thinking: Reasoning Effort as a Model-Specific API Contract},
  pdfauthor={Yeabin Moon},
  pdflang={en-US},
  colorlinks=true,
  linkcolor={black},
  filecolor={Maroon},
  citecolor={Blue},
  urlcolor={blue},
  pdfcreator={LaTeX via pandoc}}

\ifXeTeX

\fi
\ifPDFTeX
\pdftrailerid{}
\pdftrailer{/ID [ <122c3ab1b9a78bd79cb004cf3dd51d2d> <122c3ab1b9a78bd79cb004cf3dd51d2d> ]}
\fi
\ifLuaTeX
\pdfvariable trailerid {[ <122c3ab1b9a78bd79cb004cf3dd51d2d> <122c3ab1b9a78bd79cb004cf3dd51d2d> ]}
\fi

\title{The Price of Thinking: Reasoning Effort as a Model-Specific API
Contract}
\author{Yeabin Moon\\
Brandeis School of Business and Economics\\
Brandeis University\\
\href{mailto:yeabinmoon@brandeis.edu}{\texttt{yeabinmoon@brandeis.edu}}}
\date{}

\begin{document}
\maketitle

\section{Abstract}\label{abstract}

API buyers purchase a dated contract, not a model name alone: the
contract includes the requested and served model, reasoning-effort term
or its omission, output rail, service product, prompt, and price
schedule. We study the reasoning-effort term through a registered paired
contrast of Sonnet 5 with explicit high effort against the same model
with effort omitted, using 30 AIME 2026 items and five calls per item.
Every paid attempt was assigned one frozen terminal category, and
inference resampled items while retaining their repeated calls. Mean
delivered cost was \$0.01031 per call higher under the explicit-high
contract than under the omitted contract {[}+\$0.00204, +\$0.01974{]}.
The corresponding accuracy contrast was +0.0133 {[}-0.0267, +0.0467{]};
we did not detect an accuracy difference, and the interval permits a
gain of up to 4.67 percentage points that this design cannot rule out.
Cost per correct answer was \$0.08665 under the high-effort contract and
\$0.07662 under the omitted contract, as registered point estimates. A
dated contract census, Models-API metadata, and preregistered
raw-response probes further documented model-specific omission
semantics, including within a provider; claims remained at documentation
grade when raw structure was indeterminate. The request registry,
parser, terminal taxonomy, statistical plan, and analysis pipeline were
frozen before outcomes were examined; the resulting claims are bounded
to the model, task, and collection date studied.

\section{Introduction}\label{introduction}

The price printed beside a language model is not the price of a
completed task. API inference cost was already stochastic when buyers
paid for variable-length completions, but reasoning models add a less
legible source of variation: hidden thinking tokens allocated before the
visible answer. Chen et al.~(2026) report that realized cost reversed
the ordering implied by listed token prices in roughly one in three of
336 model-task comparisons and that repeated calls to fixed queries
varied substantially in cost. We take those results as established
background, not as contributions of this study.

That evidence leaves a narrower question unresolved. Chen et al.~(2026)
run each model at one reasoning setting and identify the effort-cost
ablation as future work. Their unit of comparison is therefore a
configured model: changes in provider, architecture, token prices, and
reasoning mode can move together. They also analyze cost separately from
quality. Incorrect responses retain their token costs, but paid attempts
are not classified into mutually exclusive delivered outcomes that
distinguish correct answers, wrong answers, rail-exhausted no-answer
responses, other no-answer responses, and provider failures. Those
choices fit a study of model-level price reversals, yet they do not
identify what a buyer receives by changing an effort parameter while
keeping the model fixed. Nor do they show whether two request surfaces
that expose the same nominal control implement the same omission
behavior. The relevant object for that question is not the provider or
model name alone, but the contract formed by the requested and served
model, effort setting or omission, output rail, service product, prompt,
price schedule, and run date.

We ask: given that reasoning-model cost is stochastic, what changes when
a buyer changes one explicit term of the same model's API contract, and
how should the resulting delivered outcomes be priced? Our primary
design is a registered, paired contrast of Sonnet 5 with explicit high
effort against Sonnet 5 with the effort parameter omitted. Effort and
thinking are separate request fields. In the omitted cell both were
unspecified: Sonnet 5's documented defaults were high effort and
adaptive thinking. Disabling thinking requires an explicit
\texttt{thinking:\ \{"type":\ "disabled"\}} request (Exhibit 1). The
primary cells therefore differ in whether high effort is explicit or
defaulted, not in their documented effort level or thinking mode. The
comparison holds the model, prompt template, 30 AIME 2026 items, output
rail, service product, and price schedule fixed; each item receives five
calls under each contract. The full grid adds single-pass reference
cells for GPT-5.6 Terra with effort omitted and Claude Fable 5 with
effort omitted, plus a small non-contemporaneous GPT-5.4-mini low- and
high-effort bridge. Every attempted purchase receives both a
delivered-cost measurement and one terminal category under a frozen
precedence rule. A valid final answer is correct or wrong regardless of
stop state; a rail outcome requires both rail exhaustion and no valid
final answer; other no-answer and provider-failure states remain
separate. We estimate cell means and paired contrasts by resampling
items while retaining all calls for each selected item. Repeated calls
therefore reveal conditional behavior on a problem without being counted
as additional independent problems.

Mean delivered cost was \$0.01031 per call higher under the
explicit-high contract than under the omitted contract. The point
estimate of the corresponding accuracy contrast was +0.0133, but we did
not detect an accuracy difference; the registered interval permits a
gain of up to 4.67 percentage points that some buyers could value. The
result must not be read as evidence that high effort buys nothing. The
registered cost-per-correct point estimate was also higher under the
high-effort contract than under the omitted contract, but no separate
interval was registered. Three supporting contributions place that
contrast in an auditable market setting. First, a dated contract census
and preregistered raw-response probes document that omission semantics
differ across models, including within one provider, so provider-level
descriptions are too coarse. Raw structure verified some exact-model
claims, while structurally indeterminate responses left others at
documentation grade. Second, a frozen terminal-outcome taxonomy keeps
wrong and no-answer calls in the purchased outcome mixture and its cost
total instead of treating them as missing observations. Third, the
request registry, spending limits, parser, statistical rules, and
analysis pipeline were fixed before the registered outcomes were
examined. Together these elements turn an effort label into a measurable
contract comparison while keeping the empirical claim bounded to the
model, task, and collection date studied here.

\section{Related work}\label{related-work}

Cost-aware deployment research treats model selection as an economic
decision rather than a pure accuracy ranking. Chen, Zaharia, and Zou
(2023) formalize this decision in FrugalGPT as cost-constrained utility
maximization and learn cascades that route a query through a ladder of
models, stopping when a cheaper response is adequate and escalating
otherwise. In that formulation, the available services and their cost
information are inputs to the routing problem: the router chooses among
models whose per-query prices are treated as known inputs. Our setting
changes the object being chosen, not the logic of cost-aware routing. A
buyer can alter a reasoning-effort term within one model, and omission
is itself a model-specific request state. We therefore treat the
configured API contract as the candidate service. We do not claim to
have discovered a mismatch between listed and realized prices; that
finding belongs to Chen et al.~(2026), discussed below.

Cost-adjusted evaluation asks how capability comparisons change once
inference expense enters the objective. Erol et al.~(2025) define
Cost-of-Pass, an economically framed frontier metric that combines task
success with inference cost across models and strategies. Our
cost-per-correct quantity belongs to the same family: total delivered
cost is divided by the number of correct outcomes. We do not propose it
as a new metric. Instead, we apply the quantity within one commercial
model across two registered effort contracts, report it as a point
estimate, and retain every paid attempt in the cost and terminal-outcome
accounting. Cost-of-Pass supplies the decision framing; the present
study asks whether changing a request term changes the purchased outcome
under an otherwise fixed model contract rather than constructing another
cross-model frontier.

The FrugalGPT authors themselves return to this problem in \emph{The
Price Reversal Phenomenon} (Chen et al., 2026), which examines
reasoning-model costs at substantially greater model and task breadth.
They report that listed token prices frequently misorder realized
workload costs --- roughly one in three of 336 comparisons in the pinned
v2 --- and attribute the reversals to thinking-token volume on
single-turn tasks and to additional turn and context effects in agentic
settings. They also report substantial cost variation across repeated
calls to fixed queries. We take these as established results, not
contributions of this paper. Their design runs each model at a single
reasoning setting, analyzes cost separately from quality, and does not
classify paid attempts by terminal outcome. Our narrower point of
departure is the registered within-model contrast between explicit high
effort and the omitted contract, coupled to item-clustered inference and
joint accounting of delivered cost and terminal outcomes.

\section{Methods}\label{methods}

\subsection{Task and items}\label{task-and-items}

The accuracy instrument is the complete 30-item AIME 2026 cohort
distributed through the MathArena evaluation platform (Dekoninck et al.,
2026) and the pinned \texttt{MathArena/aime\_2026} revision. We preserve
the ordered measurement IDs \texttt{aime\_2026\_i\_01} through
\texttt{aime\_2026\_i\_30}. A model response is generated freely, but
the scoring target is closed-form: the final claimed answer must be an
integer from 0 through 999. The frozen parser reads the final explicit
\texttt{Answer:\ N} claim and does not fall back to an earlier
valid-looking claim when the final claim is malformed.

AIME 2026 was chosen to reduce the contamination concern observed on the
prior year's problems. The contest was administered in February 2026,
after the documented knowledge cutoffs used in the registered design,
although the gap from Anthropic's January 2026 cutoff was narrow. In the
preliminary cohort, Sonnet low answered two AIME 2025 items correctly in
about 10 billed output tokens each, too few to contain hidden reasoning.
At the same item indices in AIME 2026, billed output rose to 755 and
1,627 tokens while the visible reply remained a short final claim,
indicating substantial hidden computation. We treat this within-index
change as a retrieval-like freshness diagnostic, not as proof that AIME
2026 is uncontaminated. AIME material is licensed CC BY-NC-SA. The
release therefore contains item identifiers, configurations,
measurements, and derived outcomes, but not the problem text or
reference answers.

\subsection{Contracts and registered
cells}\label{contracts-and-registered-cells}

The unit of comparison is a contract cell: requested model, served
model, effort term or its omission, output rail, service product,
request mode, prompt template, dated price schedule, and collection
date. This definition prevents a model-specific result from being
promoted to a provider-wide claim and makes a served-model or
service-tier mismatch an observable contract deviation rather than an
ignorable implementation detail.

The main grid contains four registered cells on all 30 items. We refer
to the full contracts as Sonnet high, Sonnet omitted, Terra omitted, and
Fable omitted. The two Sonnet cells each use five calls per item; Terra
omitted and Fable omitted each use one. Repeat depth was concentrated in
the paired Sonnet contrast, where conditional within-item behavior is a
registered estimand; the single-pass cells serve as cross-model
reference points. Two earlier \texttt{gpt-5.4-mini} cells supply a
five-item descriptive bridge. Every cell uses the frozen
\texttt{aime\_frozen\_v1} prompt template and a 64,000-token output
rail.

\textbf{Table 1. Registered AIME 2026 contract cells and descriptive
bridge cells.}

\begingroup
\footnotesize
\setlength{\tabcolsep}{2.5pt}
\renewcommand{\arraystretch}{1.08}

{\def\LTcaptype{none} 
\begin{longtable}[]{@{}
  >{\raggedright\arraybackslash}p{(\linewidth - 8\tabcolsep) * \real{0.1765}}
  >{\raggedright\arraybackslash}p{(\linewidth - 8\tabcolsep) * \real{0.1765}}
  >{\raggedleft\arraybackslash}p{(\linewidth - 8\tabcolsep) * \real{0.2353}}
  >{\raggedleft\arraybackslash}p{(\linewidth - 8\tabcolsep) * \real{0.2353}}
  >{\raggedright\arraybackslash}p{(\linewidth - 8\tabcolsep) * \real{0.1765}}@{}}
\toprule\noalign{}
\begin{minipage}[b]{\linewidth}\raggedright
Cell
\end{minipage} & \begin{minipage}[b]{\linewidth}\raggedright
Requested model and effort term
\end{minipage} & \begin{minipage}[b]{\linewidth}\raggedleft
Items x calls per item
\end{minipage} & \begin{minipage}[b]{\linewidth}\raggedleft
Calls
\end{minipage} & \begin{minipage}[b]{\linewidth}\raggedright
Collection
\end{minipage} \\
\midrule\noalign{}
\endhead
\bottomrule\noalign{}
\endlastfoot
Sonnet high & \texttt{claude-sonnet-5}, explicit \texttt{high} & 30 x 5
& 150 & 2026-07-18 \\
Sonnet omitted & \texttt{claude-sonnet-5}, effort omitted & 30 x 5 & 150
& 2026-07-18 \\
Terra omitted & \texttt{gpt-5.6-terra}, effort omitted & 30 x 1 & 30 &
2026-07-18 \\
Fable omitted & \texttt{claude-fable-5}, effort omitted & 30 x 1 & 30 &
2026-07-18 \\
Mini low bridge & \texttt{gpt-5.4-mini}, explicit \texttt{low} & 5 x 1 &
5 & 2026-07-09 \\
Mini high bridge & \texttt{gpt-5.4-mini}, explicit \texttt{high} & 5 x 1
& 5 & 2026-07-09 \\
\end{longtable}
}

\endgroup

The 360 main-grid calls were made in the single registered session
\texttt{\detokenize{aime2026_m}\allowbreak{}\detokenize{ain_202607}\allowbreak{}\detokenize{18_01}}
using non-streaming real-time requests and no automatic retries. Each
call was issued as a separate, stateless API request: no conversation
history or earlier response was supplied, and the study passed no state
from one call to the next. Repeats therefore sampled the same frozen
request contract rather than a conversation in which the model had
already seen the item. Dispatch order was fixed rather than randomized:
within all 150 Sonnet item/repeat pairs, the high call preceded the
omitted call in the queue. The first 20 pairs ran strictly sequentially;
the remaining 130 used bounded concurrency, with at most three in-flight
requests per provider, under which completion order could differ. Every
successful main-grid response reported the exact requested model ID. The
bridge alias resolved to \texttt{gpt-5.4-mini-2026-03-17}; its five
shared items were indices 1, 5, 8, 12, and 15. Because those bridge
calls are non-contemporaneous and cover only five items, their
comparisons remain descriptive.

\subsection{Dated contract census and structural
probes}\label{dated-contract-census-and-structural-probes}

Before main-grid promotion, we froze a July 14 census of seven
current-menu model IDs using official documentation and Models-API
metadata, without inference calls. On July 16, we then ran 19
preregistered, single-item contract probes on the preselected HEADLINES
task. These probes tested request acceptance, served identity, echoed
effort, thinking-block structure, thinking-token details, and explicit
disabled-thinking requests; their task accuracy was not used to select
main-grid cells.

Evidence grades followed a fixed hierarchy. A raw echoed effort value,
positive thinking structure, or acceptance or rejection of an explicit
disabled-thinking request upgraded only the contract property it
directly identified. By contrast, an absent thinking block together with
a missing or zero thinking-token field could not distinguish disabled
thinking from adaptive thinking that selected zero tokens. Such a result
remained indeterminate and did not upgrade documented omission behavior.

\subsection{Delivered cost and terminal
outcomes}\label{delivered-cost-and-terminal-outcomes}

Delivered cost is reconstructed from provider-reported usage under the
price schedule registered for each cell. The reconstruction uses the
provider's applicable input, cached-input, and output accounting,
including billed reasoning within output usage, and is checked against
the cost stored with each row. For the July 18 main session,
provider-level reconstructed totals were also reconciled against
authenticated console balance changes at the console's cent granularity.
This external reconciliation validates the accounting boundary without
changing any row or statistical estimate.

Every attempted call receives exactly one of five terminal categories:
\texttt{correct}, \texttt{wrong}, \texttt{no\_answer\_rail},
\texttt{no\_answer\_other}, or \texttt{provider\_failure}.
Classification follows a fixed precedence. A valid final claim is
correct or wrong regardless of the provider stop state. A call is
\texttt{no\_answer\_rail} only when it exhausts its registered output
rail and contains no valid final claim; a successful response without a
valid claim and without rail exhaustion is \texttt{no\_answer\_other}.
Transport, timeout, rate-limit, server, and rejected-request failures
enter \texttt{provider\_failure} under the registered operational policy
rather than being silently converted to wrong answers.

Raw provider responses are preserved before extraction. The analysis
decodes those artifacts, extracts visible text again, applies the frozen
parser, and combines the fresh parse with the normalized provider stop
state. Consequently, a local extraction exception cannot turn a
preserved, billed response into a provider failure. Wrong and no-answer
calls remain in the attempt count, terminal-outcome mixture, and total
delivered cost because they are purchased outcomes rather than missing
observations.

\subsection{Statistical plan}\label{statistical-plan}

The task item is the independent unit. For a repeated cell, calls are
first averaged within item; cell means then weight the 30 item-level
values equally. Paired contrasts are formed within shared items before
averaging. The primary uncertainty procedure is an item-clustered
percentile bootstrap: each draw samples 30 items with replacement and
carries every call and compared cell belonging to a selected item. We
use 10,000 draws, analysis seed \texttt{20260713}, and percentile 95\%
intervals. Repeats therefore estimate conditional behavior on a problem
but never convert 30 items into 150 independent problems.

The nominal 95\% level describes repeated use of the procedure, to the
extent that the bootstrap approximation holds with 30 independent items;
it is not a posterior probability that a realized interval contains the
estimand. At this sample size, percentile coverage can be imperfect or
discrete, especially for sparse outcomes or rare extreme costs absent
from the observed cohort. The bootstrap redistributes observed item
clusters but cannot create an event that no observed item contains.
Counts and item denominators therefore accompany sparse outcomes, and a
zero-event interval that collapses to zero is not interpreted as an
informative population upper bound.

For unrepeated binary proportions, we additionally report two-sided
exact 95\% Clopper-Pearson intervals. This rule applies to the Terra and
Fable main cells and the two mini bridge cells; repeated-call
proportions are not treated as independent binomial trials. Sparse
outcomes are accompanied by their raw counts and item denominators.

Tail reporting was gated before collection by the number of independent
items. Median, p90, p95, and maximum are eligible with at least 20
items. A p99 requires at least 100 items, and 90\%-CVaR requires at
least 50. The 30-item main cohort is therefore eligible for the first
set but not for p99 or CVaR. We do not report those ineligible
quantities, even though software could calculate sample analogues,
because the refusal was part of the registered analysis rather than a
reaction to the observed tail.

In preliminary collection, repeated Sonnet high calls on one AIME item
mixed cheap wrong answers with rail-censored no-answer responses,
motivating the registered rail-utilization measures. Before main
collection, we fixed the following preregistered conservative gate: if
rail outcomes were absent from the main grid, the tail claim would
narrow to the preliminary existence evidence and registered upper
bounds; if they recurred, we would describe their observed frequency
without extrapolating beyond supported quantiles.

\subsection{Governance and
reproducibility}\label{governance-and-reproducibility}

The prompt, parser, terminal taxonomy, statistical rules, contract
registry, spending ceilings, and executable analysis pipeline were
frozen before any registered outcome was inspected. Raw session and
analysis artifacts are immutable and SHA-256-pinned; the integrity
validator fails on missing, extra, truncated, or changed registered
files. Code development for the final analysis used synthetic fixtures,
and the first successful production invocation was the first read of the
main-session correctness fields. Any later deviation must be recorded as
a dated amendment stating whether the affected outcomes had already been
observed; unregistered statistics do not enter the confirmatory results.

\section{Results}\label{results}

\subsection{Primary Sonnet contract
contrast}\label{primary-sonnet-contract-contrast}

Mean delivered cost was \$0.01031 per call higher under the
explicit-high contract than under the omitted contract {[}+\$0.00204,
+\$0.01974{]}, while the accuracy contrast was +0.0133 {[}-0.0267,
+0.0467{]}. Cost per correct answer was \$0.08665 under Sonnet high and
\$0.07662 under Sonnet omitted; these are point estimates without
separately registered intervals (Table 2).

\clearpage

\textbf{Table 2. Accuracy, delivered cost, and observed cost tails by
registered AIME 2026 contract.}

\begingroup
\scriptsize
\setlength{\tabcolsep}{2.5pt}
\renewcommand{\arraystretch}{1.08}

{\def\LTcaptype{none} 
\begin{longtable}[]{@{}
  >{\raggedright\arraybackslash}p{(\linewidth - 10\tabcolsep) * \real{0.1304}}
  >{\raggedleft\arraybackslash}p{(\linewidth - 10\tabcolsep) * \real{0.1739}}
  >{\raggedleft\arraybackslash}p{(\linewidth - 10\tabcolsep) * \real{0.1739}}
  >{\raggedleft\arraybackslash}p{(\linewidth - 10\tabcolsep) * \real{0.1739}}
  >{\raggedleft\arraybackslash}p{(\linewidth - 10\tabcolsep) * \real{0.1739}}
  >{\raggedleft\arraybackslash}p{(\linewidth - 10\tabcolsep) * \real{0.1739}}@{}}
\toprule\noalign{}
\begin{minipage}[b]{\linewidth}\raggedright
Contract
\end{minipage} & \begin{minipage}[b]{\linewidth}\raggedleft
Calls (items x repeats)
\end{minipage} & \begin{minipage}[b]{\linewidth}\raggedleft
Accuracy (correct/calls), 95\% CI
\end{minipage} & \begin{minipage}[b]{\linewidth}\raggedleft
Mean cost/call, 95\% CI
\end{minipage} & \begin{minipage}[b]{\linewidth}\raggedleft
Cost/correct
\end{minipage} & \begin{minipage}[b]{\linewidth}\raggedleft
Cost tail: median / p95 / max
\end{minipage} \\
\midrule\noalign{}
\endhead
\bottomrule\noalign{}
\endlastfoot
GPT-5.6 Terra, effort omitted & 30 (30 x 1) & 0.933 (28/30) {[}0.779,
0.992{]} & \$0.02319 {[}\$0.01745, \$0.02992{]} & \$0.02485 & \$0.01801
/ \$0.05093 / \$0.08884 \\
Claude Sonnet 5, effort omitted & 150 (30 x 5) & 0.913 (137/150)
{[}0.813, 0.987{]} & \$0.06998 {[}\$0.04427, \$0.10062{]} & \$0.07662 &
\$0.04093 / \$0.24428 / \$0.53290 \\
Claude Sonnet 5, explicit high effort & 150 (30 x 5) & 0.927 (139/150)
{[}0.833, 1.000{]} & \$0.08029 {[}\$0.05107, \$0.11587{]} & \$0.08665 &
\$0.04169 / \$0.31741 / \$0.58653 \\
Claude Fable 5, effort omitted & 30 (30 x 1) & 1.000 (30/30) {[}0.884,
1.000{]} & \$0.21582 {[}\$0.11158, \$0.36711{]} & \$0.21582 & \$0.10466
/ \$0.72662 / \$1.98876 \\
GPT-5.4-mini, low (July 9 bridge; descriptive) & 5 (5 x 1) & 0.800 (4/5)
{[}0.284, 0.995{]} & \$0.01006 {[}\$0.00401, \$0.01657{]} & \$0.01257 &
Not reported (5 items) \\
GPT-5.4-mini, high (July 9 bridge; descriptive) & 5 (5 x 1) & 0.800
(4/5) {[}0.284, 0.995{]} & \$0.04082 {[}\$0.00511, \$0.10396{]} &
\$0.05102 & Not reported (5 items) \\
\end{longtable}
}

\endgroup

Notes: Accuracy is shown as the observed call proportion with the raw
correct/call count. Accuracy intervals for the repeated Sonnet cells use
the registered item-clustered bootstrap, which resamples items while
retaining all five calls; exact Clopper-Pearson intervals are shown for
the single-pass and bridge cells. Repeated calls do not increase the
independent-item count. Mean-cost intervals use the registered
item-clustered bootstrap. Cost per correct is the frozen total delivered
cost divided by the observed correct count and is copied from the
registered statistics rather than recomputed for this table. Cost tails
report the registered call-level median, p95, and maximum only for cells
meeting the 20-independent-item eligibility gate. Bridge cells are
descriptive, non-contemporaneous anchors and do not meet that gate.
Dollar values use the dated price schedules registered for each contract
and Python fixed-point formatting applied to the stored floating-point
values. Latency is not represented.

\subsection{Dated omission and thinking-control
evidence}\label{dated-omission-and-thinking-control-evidence}

\textbf{Exhibit 1. Contract-control evidence frozen before the main-grid
outcomes were examined.}

\begin{itemize}
\tightlist
\item
  \textbf{GPT-5.6 Sol, Terra, and Luna.} Documentation assigned omitted
  effort to \texttt{medium}. Each omitted probe echoed
  \texttt{reasoning.effort=medium}, raw-verifying that behavior for
  these exact model IDs and the probe date
  (\texttt{structural\_raw\_response}).
\item
  \textbf{Claude Fable 5.} Documentation assigned omitted effort to
  \texttt{high} and described adaptive thinking as always on. The
  omitted probe contained positive thinking structure, while an explicit
  disabled-thinking request returned an HTTP 400 error naming the
  unsupported mode. These raw artifacts verify realized thinking on the
  omitted probe and the absence of an off-switch, not a general
  thinking-token level (\texttt{structural\_raw\_response};
  \texttt{structural\_request\_error}).
\item
  \textbf{Claude Opus 4.8.} Documentation assigned omitted effort to
  \texttt{high} but described omitted thinking as off. The omitted probe
  contained no thinking block and reported zero thinking tokens, which
  was consistent with but did not raw-verify that behavior. A
  disabled-thinking-plus-high request was accepted. Omission therefore
  remained documentation-grade with indeterminate raw evidence
  (\texttt{\detokenize{documentat}\allowbreak{}\detokenize{ion_plus_i}\allowbreak{}\detokenize{ndetermina}\allowbreak{}\detokenize{te_raw}};
  \texttt{\detokenize{request_su}\allowbreak{}\detokenize{rface_raw_}\allowbreak{}\detokenize{response}}
  for the accepted disabled request).
\item
  \textbf{Claude Sonnet 5.} Documentation assigned omitted effort to
  \texttt{high} and described omitted thinking as adaptive. The omitted
  probe contained no thinking block and reported zero thinking tokens,
  which could not distinguish adaptive-zero from disabled thinking. A
  disabled-thinking-plus-high request was accepted. Omission therefore
  remained documentation-grade with indeterminate raw evidence
  (\texttt{\detokenize{documentat}\allowbreak{}\detokenize{ion_plus_i}\allowbreak{}\detokenize{ndetermina}\allowbreak{}\detokenize{te_raw}};
  \texttt{\detokenize{request_su}\allowbreak{}\detokenize{rface_raw_}\allowbreak{}\detokenize{response}}
  for the accepted disabled request).
\item
  \textbf{Claude Haiku 4.5.} Documentation and Models-API metadata
  exposed no effort control. The probe served the pinned model snapshot
  without thinking structure, providing consistency evidence but not
  testing an unrequested effort parameter
  (\texttt{\detokenize{metadata_p}\allowbreak{}\detokenize{lus_struct}\allowbreak{}\detokenize{ural_raw_r}\allowbreak{}\detokenize{esponse}}).
\end{itemize}

The exhibit separates documented behavior from structural raw-response
evidence rather than treating every successful probe as verification.
The dated source record is preserved in the dated census, contract
registry, probe expectation, and probe result records.

Descriptively, the registered grid rows bear on the same question: 140
of the 150 Sonnet omitted calls delivered positive reasoning-token
counts (explicit high: 141 of 150), with positive-call medians of
3,133.5 versus 3,359 tokens. Positive counts show that thinking occurred
in those 140 omitted calls and therefore that omission did not act as
disabled thinking in them. The two medians are post hoc descriptive
quantities and do not establish equivalence between request forms or a
causal effect of explicitly setting high. None of these values is a
registered statistic.

\subsection{Delivered-cost
distributions}\label{delivered-cost-distributions}

Figure 1 displays all 360 registered calls by contract and terminal
category. Median delivered cost was \$0.01801 for Terra omitted,
\$0.04093 for Sonnet omitted, \$0.04169 for Sonnet high, and \$0.10466
for Fable omitted. The two Sonnet contracts had similar medians even
though their individual calls occupied visibly different parts of the
cost range. Both also contained a low-cost cluster that included
unsuccessful outcomes; we treat that cluster qualitatively rather than
imposing an unregistered cost threshold or reporting a post hoc share.
Each dot in the figure remains a call nested within an item. The display
therefore exposes call-level overlap without treating the dots as
independent evidence.

\begin{figure}[htbp]
\centering
\includegraphics[width=0.96\linewidth,height=0.43\textheight,keepaspectratio]{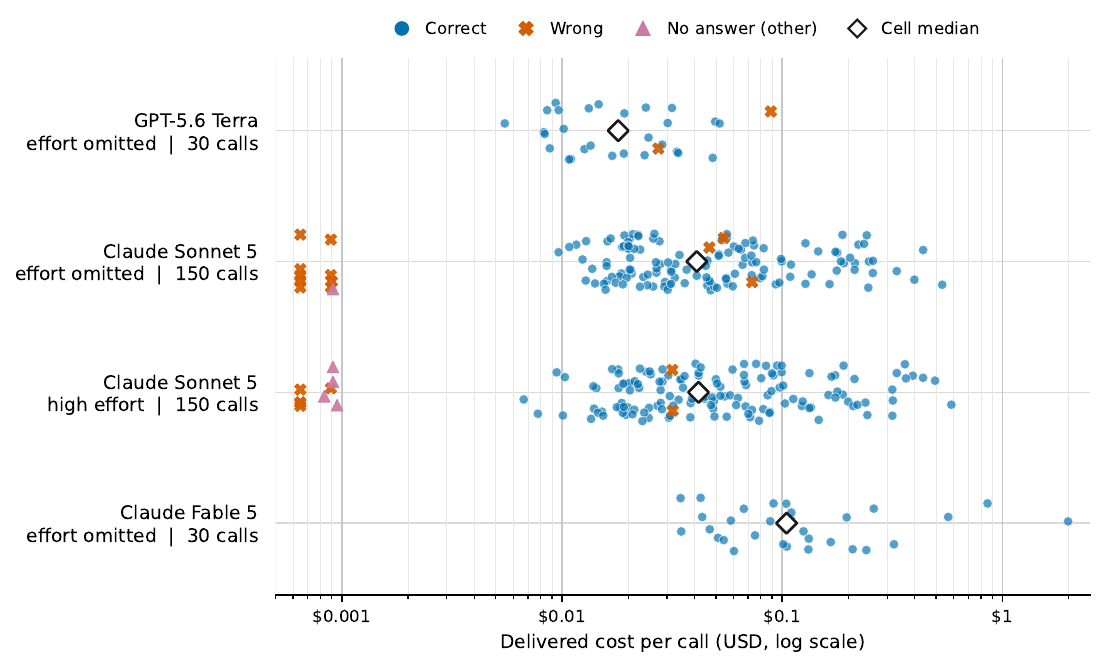}
\par\smallskip
\begin{minipage}{0.96\linewidth}
\small

\textbf{Figure 1. Delivered cost distributions under four registered
AIME 2026 contracts.} Each point represents one provider call. The two
Sonnet cells contain five repeated calls on each of 30 items; Terra and
Fable contain one call per item. The task item is the inferential unit,
and repeated calls remain grouped within item rather than being treated
as additional independent observations. The figure itself contains no
uncertainty intervals. Color and shape indicate the terminal outcome,
and open diamonds mark the registered cell medians. Delivered cost is
shown on a logarithmic scale. Latency is not represented.

\end{minipage}
\end{figure}

\subsection{Accuracy-cost frontier and historical
bridge}\label{accuracy-cost-frontier-and-historical-bridge}

Figure 2 places the four main cells on the registered accuracy-cost
plane. Terra omitted answered 28/30 items correctly, with exact interval
{[}0.779, 0.992{]}, at mean delivered cost \$0.02319 {[}\$0.01745,
\$0.02992{]}. Sonnet omitted had accuracy 0.913 {[}0.813, 0.987{]} and
mean cost \$0.06998 {[}\$0.04427, \$0.10062{]}; Sonnet high had accuracy
0.927 {[}0.833, 1.000{]} and mean cost \$0.08029 {[}\$0.05107,
\$0.11587{]}. For Fable omitted, no failure was observed in 30 items;
its exact accuracy interval was {[}0.884, 1.000{]}, and mean cost was
\$0.21582 {[}\$0.11158, \$0.36711{]}. Because all 30 calls were correct,
its cost-per-correct point estimate was numerically identical to its
mean cost.

The two July 9 mini bridge cells are descriptive and
non-contemporaneous. Low and high each answered 4/5 items correctly,
with exact interval {[}0.284, 0.995{]}; their mean costs were \$0.01006
{[}\$0.00401, \$0.01657{]} and \$0.04082 {[}\$0.00511, \$0.10396{]},
respectively.

\begin{figure}[htbp]
\centering
\includegraphics[width=0.96\linewidth,height=0.43\textheight,keepaspectratio]{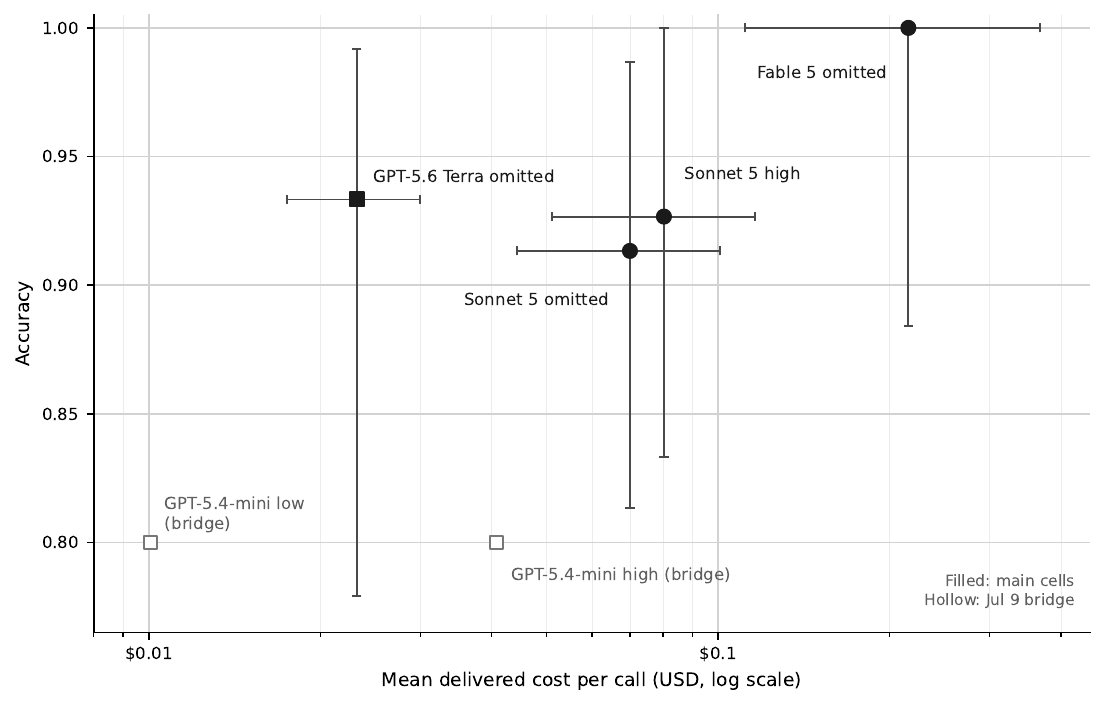}
\par\smallskip
\begin{minipage}{0.96\linewidth}
\small

\textbf{Figure 2. Accuracy and mean delivered cost under registered AIME
2026 contracts.} Points locate mean delivered cost per registered call
against accuracy. For the two repeated Sonnet cells, vertical intervals
are registered 95\% item-clustered bootstrap intervals. For the
single-pass Terra and Fable cells, vertical intervals are two-sided 95\%
exact Clopper-Pearson intervals. Horizontal intervals are registered
95\% item-clustered bootstrap intervals for mean delivered cost. The two
smaller hollow points are the non-contemporaneous five-item GPT-5.4-mini
bridge cells collected on July 9 and are shown without interval bars;
each observed 4/5 correct, with exact 95\% interval {[}0.284, 0.995{]}.
Fable produced 30/30 observed correct responses, with exact 95\%
interval {[}0.884, 1.000{]}; this is not evidence of population accuracy
equal to 100\%. Mean delivered cost is shown on a logarithmic scale.
Latency is not represented.

\end{minipage}
\end{figure}

\subsection{Within-item variation}\label{within-item-variation}

Figure 3 compares the registered minimum, mean, and maximum delivered
cost for each item in the repeated Sonnet cells; these spans are
descriptive, not uncertainty intervals. Sonnet high produced more than
one terminal category on items 2, 11, and 15, and Sonnet omitted did so
on items 15, 17, 18, and 25. Item 11 was selected after outcome
inspection as a descriptive illustration; the figure displays all 30
items, and no population-rate claim is based on that selection. At the
stored six-decimal precision, Sonnet high on item 11 combined wrong and
correct outcomes across a cost span from \$0.000648 to \$0.391888; under
Sonnet omitted, all five calls were wrong and delivered cost was
degenerate at \$0.000648. The preliminary item 15 mixture paired cheap
wrong answers with rail-censored no-answers (Methods), a different
second branch from item 11's pairing of cheap wrong answers with one
costly correct computation, so item 11 is not a replication.

\begin{figure}[htbp]
\centering
\includegraphics[width=0.96\linewidth,height=0.57\textheight,keepaspectratio]{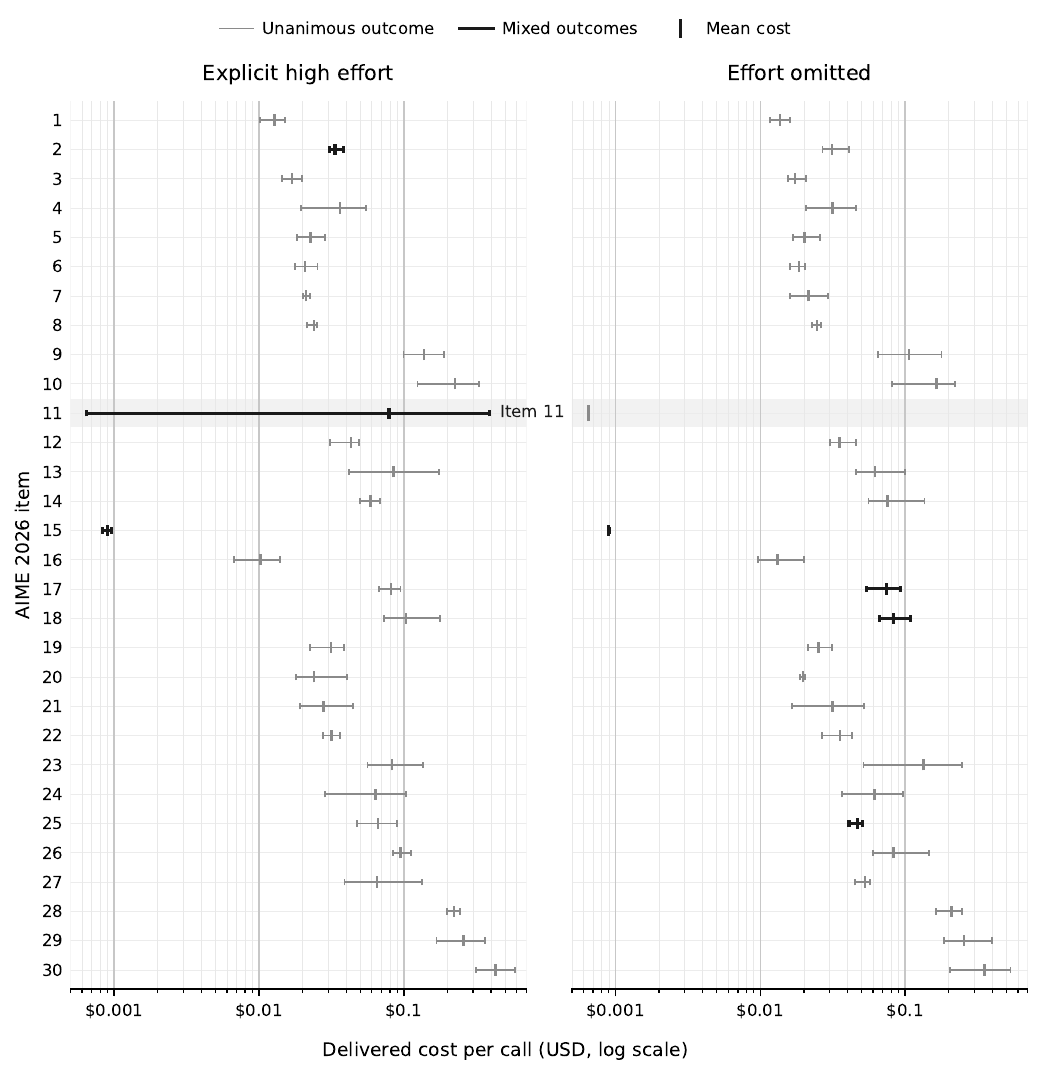}
\par\smallskip
\begin{minipage}{0.96\linewidth}
\small

\textbf{Figure 3. Within-item delivered-cost spans under repeated Sonnet
5 contracts.} Each horizontal line spans the minimum to maximum
delivered cost across the five registered calls on one AIME 2026 item;
end caps mark those endpoints and the interior tick marks the mean.
These spans are descriptive ranges, not uncertainty intervals.
Zero-width spans are shown as a single mean marker. Darker, heavier
lines identify items that produced more than one terminal-outcome
category across the five repeats; gray lines identify unanimous
outcomes. Sonnet omitted had four mixed-outcome items, compared with
three under Sonnet high. Endpoint values are printed at their stored
six-decimal precision: on item 11, Sonnet high ranged from \$0.000648 to
\$0.391888, while Sonnet omitted produced five identical wrong answers
at \$0.000648. Item 11 was selected after outcome inspection as a
descriptive illustration; all 30 items remain visible. Costs are shown
on a logarithmic scale. Latency is not represented.

\end{minipage}
\end{figure}

\subsection{Rail utilization and preregistered conservative gate
resolution}\label{rail-utilization-and-preregistered-conservative-gate-resolution}

No main-grid call ended in \texttt{no\_answer\_rail}, and none of the
360 calls reached 99\% rail utilization. For Sonnet high, 7/150 calls
reached at least 50\% utilization, with item-bootstrap interval
{[}0.0067, 0.1000{]}; 1/150 reached at least 90\%, with interval
{[}0.0000, 0.0200{]}; and 0/150 reached 99\%. Sonnet omitted had 4/150
calls at 50\% {[}0.0000, 0.0733{]} and 0/150 at both 90\% and 99\%. For
these three zero-event repeated-cell estimates, all registered bootstrap
draws were zero, mechanically yielding {[}0.0000, 0.0000{]}. They are
not informative population upper bounds.

Fable omitted had 1/30 calls at 50\%, with exact interval {[}0.0008,
0.1722{]}, and 0/30 at both higher thresholds, each with exact interval
{[}0.0000, 0.1157{]}. Terra omitted had 0/30 calls at every threshold,
also with exact interval {[}0.0000, 0.1157{]} at each. The preliminary
full-rail mixture therefore did not generalize, resolving the
preregistered conservative gate (Methods). As specified in Methods, p99
and CVaR remain unreported. Rail utilization measures delivered output
at the registered boundary and does not identify the unconstrained
resources a censored call would have required.

\subsection{Latency in the first-20-item sequential subsample
(descriptive)}\label{latency-in-the-first-20-item-sequential-subsample-descriptive}

Latency results apply only to the registered first-20-item sequential
subsample and are descriptive. Median latency was 7.8 s under Terra
omitted, 17.1 s under Sonnet omitted, 16.7 s under Sonnet high, and 17.8
s under Fable omitted. Fable's eligible upper-tail summaries were p95
226.1 s and maximum 482.9 s. These values describe the registered
subsample and are not interpreted as causal effects of effort. They
exclude the later bounded-concurrency phase, and the same scope
restriction will accompany any latency table or figure.

\section{Discussion}\label{discussion}

The primary result is a within-model contract contrast, not a provider
ranking. The registered comparison detected higher delivered cost under
the explicit-high contract but did not detect an accuracy difference.
This result must not be read as showing that high effort buys nothing:
the accuracy interval permits a gain of up to 4.67 percentage points,
which a buyer could value at the observed premium. The registered
cost-per-correct point estimate was also higher under Sonnet high, but
no separate interval was registered. The narrower implication is that
Sonnet high and Sonnet omitted are distinct request contracts within one
model. This extends cost-aware routing's model-level price input with a
controlled contract choice; it does not claim to discover stochastic
inference cost, fixed-query variation, price reversal, or the need for
distributional routing, which Chen et al.~(2026) report at much greater
breadth.

Repeated calls clarify what the two Sonnet contracts delivered on
particular items, without establishing how prevalent those behaviors
are. In the preliminary item 15 case, one fixed cell paired cheap wrong
answers with censored full-rail no-answers. In the registered item 11
case, high effort paired cheap wrong answers with one costly correct
computation, whereas Sonnet omitted produced five identical cheap wrong
answers. These observations establish two item-specific instances of
qualitatively different terminal outcomes under fixed contracts; they do
not establish a population rate. Item 11 is also not a replication of
item 15 because its second branch was a correct completion rather than a
rail-censored no-answer. The main grid contained no
\texttt{no\_answer\_rail} outcome and no call at or above 99\% rail
utilization, so the preregistered conservative gate fired. The rail
result therefore narrows to the preliminary existence case, the
registered zero counts, and applicable exact single-pass bounds; the
degenerate repeated-cell bootstrap intervals provide no informative
upper bound. Rail exhaustion also censors latent demand: the study
observes delivered cost and the missing answer, not the additional
computation that would have completed the response. Consistent with the
registered item-count gates, we make no p99 or CVaR claim.

The observed failure modes also differed in buyer legibility across the
exact main-grid cells. Terra's unsuccessful calls ended in visible wrong
answers, while the Sonnet cells included no-text no-answer outcomes in
addition to wrong answers. This is a descriptive distinction: it
concerns what a buyer can observe after purchasing these particular
dated contracts. It does not establish a provider-wide failure regime,
population prevalence, or a causal difference attributable to the
providers. Any use of this contrast must remain conditional on the
registered task, prompt, rails, service products, prices, and collection
date.

Across the four main cells, the descriptive cost-per-correct point
estimates placed Terra omitted lowest, followed by Sonnet omitted,
Sonnet high, and Fable omitted. This ordering does not imply dominance:
no separate cost-per-correct intervals were registered, and the
underlying cost and accuracy intervals overlap. The five-item July 9
bridge supplies historical context rather than additional confirmatory
evidence. The three bridge-linked cells shared the same observed
accuracy pattern, while Terra's mean delivered cost lay between the
non-contemporaneous GPT-5.4-mini low- and high-effort means. That dated
comparison echoes the point-cost decision variable used in the FrugalGPT
lineage, but its five-item denominator and non-contemporaneous
collection preclude equivalence, tail, latency, or population claims.

Two supporting features make the contract comparison auditable without
enlarging its scope. First, the July 2026 model menu assigned different
documented and raw-verified meanings to omission and thinking control,
including variation within one provider. Omission must therefore be
recorded at the exact model-contract level rather than treated as a
provider-wide default. Evidence grades remain asymmetric: positive
thinking structure verifies realized thinking, while its absence on an
easy response cannot distinguish disabled thinking from adaptive
thinking that selected zero tokens. Second, the parser, outcome
taxonomy, spending ceiling, manifest, and numerical pipeline were frozen
before outcome inspection, with later corrections recorded as dated
derived artifacts. This governance supports the credibility and
auditability of the estimates; it does not expand their empirical reach
or restore a first-discovery claim about cost stochasticity or
distributions.

Future work should treat effort choice and model choice as dimensions of
one registered measurement program. A wider design could cross an effort
ladder within a model with a contemporaneous model ladder from the same
provider menu, allowing the delivered cost of raising effort to be
compared with the delivered cost of switching models. Those
contract-level measurements could also enter effort-aware cascade
policies, alongside a deployable confidence or stopping signal, so
routing selects among configured contracts rather than model names
alone. Because prices, model menus, and omission semantics change, such
comparisons would require remeasurement when the available menu changes.

\section{Limitations}\label{limitations}

The AIME 2025-to-2026 within-index comparison is a freshness diagnostic,
not proof that AIME 2026 was uncontaminated. The instant-correct pattern
observed at matched 2025 indices disappeared in the 2026 preliminary
rows, but the approximately one-month gap from Anthropic's stated cutoff
was narrow. Cutoff metadata, possible benchmark leakage, and later model
updates remain uncertain. The main cohort also contains only 30 curated
items from one competition-mathematics task family. It is not a
probability sample from a formally defined population of model uses.
Interpretation beyond the fixed cohort therefore assumes exchangeability
with a notional population of comparable contest problems, and should
not extend to other tasks without new evidence. Observed accuracy was
high in all four main cells, leaving limited headroom for detecting
gains; nevertheless, the registered accuracy interval permits an
improvement of up to 4.67 percentage points under the explicit-high
contract.

The repetition design is asymmetric. Sonnet high and Sonnet omitted were
repeated five times per item, but Terra omitted and Fable omitted were
observed once per item. The latter cells support cross-item contract
summaries but cannot identify within-item repeat variability or
conditional outcome composition. Thirty independent items also limit
tail resolution. The registered plan permits the median, p90, p95, and
maximum, but refuses to estimate p99 with fewer than 100 items or
90\%-CVaR with fewer than 50. Repeated calls do not increase those
independent-item counts, and numerical computability does not make the
ineligible statistics informative.

Latency has a narrower scope than the other call-level measurements. Its
reported summaries apply only to the first 20 item-distinct sequential
calls in each main cell, not to the full grid or the later
bounded-concurrency phase. They are descriptive and do not identify a
causal effect of effort on response time. Because the Sonnet cells were
dispatched in a fixed high-before-omitted order rather than randomized
or counterbalanced, the design also cannot separate request-form
differences from systematic dispatch-position or short-run service
effects. We therefore interpret the registered contrast as descriptive
of the July 18 session, not as a general causal effect of explicitly
setting high. More broadly, every main-cell estimate is conditional on
that session, the served models and service products observed then, and
the price schedule in force on that date. Later model updates, routing
changes, prices, tiers, or provider policies could define different
contracts and are outside this study.

The GPT-5.4-mini bridge is a five-item, non-contemporaneous comparison
collected on July 9. It provides a dated descriptive anchor but supports
no equivalence, population accuracy, or cost- and latency-tail
inference. Finally, billed hidden computation is behaviorally measurable
but mechanistically opaque: provider records reveal its volume and
price, while its content remains unavailable. We can relate delivered
cost to the requested contract, visible output, stop state, and terminal
outcome, but cannot attribute the primary cost contrast to particular
reasoning content or provide the mechanism decomposition that Chen et
al.~(2026) report.

\section{Data, code, and spend
disclosure}\label{data-code-and-spend-disclosure}

AIME 2026 was loaded from \texttt{MathArena/aime\_2026} (MathArena,
2026) at pinned Hub revision
\texttt{\detokenize{d2de22f3c6}\allowbreak{}\detokenize{56b4f56cf8}\allowbreak{}\detokenize{9812121863}\allowbreak{}\detokenize{77d1e23bc3}}
and is licensed CC BY-NC-SA 4.0. The study release contains item
identifiers, contract configurations, row-level measurements, and
derived terminal outcomes, but no AIME problem text or reference
answers. HEADLINES (Sinha and Khandait, 2020) is licensed CC BY-NC-ND
4.0 and is not redistributed. The numerical release consists of
\texttt{release\_rows.jsonl}, \texttt{cell\_statistics.json},
\texttt{paired\_contrasts.json},
\texttt{\detokenize{within_ite}\allowbreak{}\detokenize{m_variatio}\allowbreak{}\detokenize{n.json}},
and \texttt{bridge\_statistics.json}; \texttt{artifact\_hashes.json}
supplies their SHA-256 integrity pins. The release also includes
\texttt{\detokenize{plot_figur}\allowbreak{}\detokenize{e1_cost_di}\allowbreak{}\detokenize{stribution}\allowbreak{}\detokenize{.py}},
\texttt{plot\_figure2\_frontier.py}, and
\texttt{\detokenize{plot_figur}\allowbreak{}\detokenize{e3_within_}\allowbreak{}\detokenize{item_span.}\allowbreak{}\detokenize{py}}.
Figure renders are reproducible from those pinned artifacts but are not
shipped. Raw run artifacts remain immutable; corrections and
reconciliations are stored as dated derived sidecars rather than
rewriting the source rows.

Total API expenditure through the July 18 main collection was
\$33.043144: \$31.755307 on Anthropic and \$1.287837 on OpenAI. The
total comprises \$3.303106 in precollection runs, \$0.020623 in the July
16 contract probes, \$29.711815 in the main session, and a \$0.007600
Anthropic charge from an interrupted non-streaming request. The
supplemental charge is dashboard-confirmed in
\texttt{\detokenize{results/sp}\allowbreak{}\detokenize{end_reconc}\allowbreak{}\detokenize{iliation_2}\allowbreak{}\detokenize{0260722.js}\allowbreak{}\detokenize{on}}.
Main-session row totals matched both provider dashboards at cent
granularity. The probe-day subtotal matches the immutable usage-priced
rows, but authenticated dashboard access was unavailable for an
aggregate probe-day match. Exact aggregate dashboard evidence is
likewise unavailable for the precollection period; its accounting
consists of the row-level reconstruction plus the individually confirmed
interrupted-request charge. Fifty-four error rows retain null rather
than zero cost: 50 belong to one July 7 connection-failed smoke batch,
and four are intentional validation-error probes. No usage-priced cost
was recorded for those rows; treating the rejected or failed requests as
unbilled is documentation-derived rather than request-level billing
verification.

\section{References}\label{references}

Chen, Lingjiao, Matei Zaharia, and James Zou. 2023. ``FrugalGPT: How to
Use Large Language Models While Reducing Cost and Improving
Performance.''
\href{https://arxiv.org/abs/2305.05176}{arXiv:2305.05176}.

Chen, Lingjiao, Chi Zhang, Yeye He, Ion Stoica, Matei Zaharia, and James
Zou. 2026. ``The Price Reversal Phenomenon: When Cheaper Reasoning
Models Cost More.''
\href{https://arxiv.org/abs/2603.23971v2}{arXiv:2603.23971v2}.

Dekoninck, Jasper, Nikola Jovanović, Tim Gehrunger, Kári Rögnvaldsson,
Ivo Petrov, Chenhao Sun, and Martin Vechev. 2026. ``Beyond Benchmarks:
MathArena as an Evaluation Platform for Mathematics with LLMs.''
\href{https://arxiv.org/abs/2605.00674}{arXiv:2605.00674}.

Erol, Mehmet Hamza, Batu El, Mirac Suzgun, Mert Yuksekgonul, and James
Zou. 2025. ``Cost-of-Pass: An Economic Framework for Evaluating Language
Models.'' \href{https://arxiv.org/abs/2504.13359}{arXiv:2504.13359}.

MathArena. 2026. ``AIME 2026.'' Dataset on Hugging Face, revision
\texttt{\detokenize{d2de22f3c6}\allowbreak{}\detokenize{56b4f56cf8}\allowbreak{}\detokenize{9812121863}\allowbreak{}\detokenize{77d1e23bc3}}.
\href{https://huggingface.co/datasets/MathArena/aime_2026}{MathArena/aime\_2026}.

Sinha, Ankur, and Tanmay Khandait. 2020. ``Impact of News on the
Commodity Market: Dataset and Results.''
\href{https://arxiv.org/abs/2009.04202}{arXiv:2009.04202}.

\end{document}